\documentclass{article}

\PassOptionsToPackage{numbers}{natbib}

\usepackage[final]{nips_2016}

\usepackage{amsfonts}       
\usepackage{amsmath}
\usepackage{array}
\usepackage{booktabs}       
\usepackage{enumitem}
\usepackage[T1]{fontenc}    
\usepackage{graphicx}
\usepackage{hyperref}       
\usepackage[utf8]{inputenc} 
\usepackage{microtype}      
\usepackage{movie15}
\usepackage{nicefrac}       
\usepackage{subcaption}
\usepackage{url}            

\newcolumntype{L}[1]{>{\raggedright\let\newline\\\arraybackslash\hspace{0pt}}m{#1}}
\newcolumntype{C}[1]{>{\centering\let\newline\\\arraybackslash\hspace{0pt}}m{#1}}
\newcolumntype{R}[1]{>{\raggedleft\let\newline\\\arraybackslash\hspace{0pt}}m{#1}}

\title{Optimizing and Visualizing Deep Learning for Benign/Malignant Classification in Breast Tumors}

\author{
  Darvin Yi\\
  Biomedical Data Science\\
  Stanford University\\
  Stanford, CA 94305 \\
  \texttt{darvinyi@stanford.edu} \\
  \And
  Rebecca Lynn Sawyer \\
  Biomedical Data Science\\
  Stanford University\\
  Stanford, CA 94305 \\
  \And
  David Cohn III \\
  Biomedical Data Science\\
  Stanford University\\
  Stanford, CA 94305 \\
  \And
  Jared Dunnmon \\
  Mechanical Engineering\\
  Stanford University\\
  Stanford, CA 94305 \\
  \And
  Carson Lam \\
  Biomedical Data Science\\
  Stanford University\\
  Stanford, CA 94305 \\
  \And
  Xuerong Xiao \\
  Electrical Engineering\\
  Stanford University\\
  Stanford, CA 94305 \\
  \And
  Daniel Rubin \\
  Biomedical Data Science, Radiology\\
  Stanford University\\
  Stanford, CA 94305 \\
}

\begin{document}

\maketitle

\begin{abstract}
  Breast cancer has the highest incidence and second highest mortality rate for women in the US.  Our study aims to utilize deep learning for benign/malignant classification of mammogram tumors using a subset of cases from the Digital Database of Screening Mammography (DDSM).  Though it was a small dataset from the view of Deep Learning  ($\sim 1000$ patients), we show that currently state of the art architectures of deep learning can find a robust signal, even when trained from scratch.  Using convolutional neural networks (CNNs), we are able to achieve an accuracy of 85\% and an ROC AUC of 0.91, while leading hand-crafted feature based methods are only able to achieve an accuracy of 71\%.  We investigate an amalgamation of architectures to show that our best result is reached with an ensemble of the lightweight GoogLe Nets tasked with interpreting both the coronal caudal view and the mediolateral oblique view, simply averaging the probability scores of both views to make the final prediction.  In addition, we have created a novel method to visualize what features the neural network detects for the benign/malignant classification, and have correlated those features with well known radiological features, such as spiculation.  Our algorithm significantly improves existing classification methods for mammography lesions and identifies features that correlate with established clinical markers.
\end{abstract}

\section{Introduction}\label{sec:Introduction}

Millions of mammograms are performed every year with the goal of enabling improved treatment outcomes and longer survival times for breast cancer patients via early detection \citep{rao2010widely}. Mammogram interpretation requires the expertise of a highly trained radiologist, which can be time-consuming and prone to interpretation variability, interpreter fatigue, and various interpretation errors, including both false negatives and false positives. Due to these limitations, substantial interest has arisen in the potential of Computer assisted diagnosis/detection (CAD) tools to automate medical image analysis \citep{de2016machine, wang2012machine}.

\indent CAD tools have historically relied on manually curated features.  Though hand crafted features show distinct promise in mammography classification challenges \citep{giger2008anniversary}\cite{chu2015}\cite{bekker2016}, they suffer from a variety of drawbacks: (1) hand crafted features tend to be domain (or even subtype) specific, and the process of feature design and extraction can be tedious, difficult, and non-generalizable \citep{gutman2013mr}; (2) manual feature extraction methods bias computational analysis tools towards conclusions that radiologists already make; and (3) multi-modal datasets are only integrated in ad-hoc fashion.  In light of these limitations, and in combination with the recent explosion in computational power, data-driven machine learning techniques have become of great interest in improving CAD capabilities \citep{de2016machine}.  A key enabling advance has been the advent of “deep learning,” which allows computers to build predictive algorithms based on the features that are found to best explain observed data in a generative fashion.  Fundamentally, instead of relying on human design to determine task-relevant features, supervised deep learning models derive predictive transformations via a data-based mathematical optimization process that penalizes inconsistencies between model output and ground truth.  Further, in addition to demonstrating super-human performance on such tasks as metastasis detection \citep{wang2016deep} and mammography classification \citep{kooi2017large}, deep learning enables the potential integration of heterogeneous datasets including multiple imaging views.  While deep learning models have exhibited encouraging performance in breast cancer imaging analysis tasks, these models are often treated as black boxes, and relating high-level features to clinically relevant phenomena has proven difficult.

\indent Such advances in deep learning have already had an impact on the field of breast cancer CAD in a variety of domains.  In recent work, for instance, \cite{kooi2017large} use a network similar to that described by \cite{simonyan2014very} along with targeted data augmentation to achieve an AUC of over 0.93 on mammography classification on a dataset of over 45,000 digital mammography images.  Comparison to traditional CAD techniques shows that the CNN-based method alone outperforms the traditional CAD in terms of AUC, but can also be improved by incorporating hand-specified features to deal with edge cases and specific phenomena that cause false positives or negatives.  Importantly, these results demonstrate that CNN-based systems show great promise as a second reader in a radiological workflow -- however, substantial work remains to improve these models to the point that a CAD system could accomplish independent detection.  

\indent As another example of the potential for deep learning to enable improved CAD for breast cancer, the Camelyon Grand Challenge focuses on the task of metastatic breast cancer detection from sentinel lymph node biopsies whole slide images. After basic preprocessing of the data, \cite{wang2016deep} used a GoogLeNet architecture to generate probability maps indicating the likelihood of metastasis for each pixel. This result was used not only to determine whether a slide was indicative of metastasis, but also to label locations on the slide that were most indicative of this phenomenon using both deep learning output and a random forest technique leveraging hand-designed features. Importantly, \cite{wang2016deep} report an AUC of 0.925 on the slide classification task while a pathologist reported an AUC of 0.966 -- however, with the aid of the model's predictions, the pathologist AUC was increased to 0.995. This result is particularly powerful because it suggests that deep learning approaches for medical image analysis may increase practitioner accuracy, as the errors made by by the human and machine seem to be fundamentally independent \citep{wang2016deep}.

\indent However, while deep learning models have exhibited encouraging performance in breast cancer imaging analysis tasks, these models are often treated as black boxes, and relating high-level features to clinically-relevant phenomena has proven difficult. Fortunately, recent work in the field of image classification has shown promise in relating convolutional features to the underlying structure and performance of deep learning models. \citep{simonyan2013deep}, for instance, present two relevant visualization techniques. The first generates an image that maximizes a given class score via back-propagation, starting from a blank image. The second generates a saliency map for a given image and class by performing back-propagation for a specific class score with respect to the input image \cite{simonyan2013deep}. The DeepDream algorithm, introduced in 2015, has also been used to determine what patterns are present in an image by enhancing learned features at different layers in the network. Specifically, this algorithm uses a model pretrained to discriminate between images on some data set and performs gradient ascent to maximize the L2 norm of activations at a given layer \cite{mordvintsev2015inceptionism}.  In the context of medical imaging, it becomes immensely valuable to be enable model interpretability and visualization, as in addition to improving performance, understanding why a deep learning model classifies an image in a certain way may well enable novel scientific discoveries while yielding more accurate diagnoses.  We are therefore interested not only in performing well on an important task such as mammography classification, but also in visualizing higher-level features that may be clinically relevant. Thus, in the current work we also present a visualization technique designed to understand the clinical and biological relevance of high-level features learned by the model.

\indent In this paper, we make several distinct contributions that relate recent advances in computer vision to applications in mammography classification: (1) demonstrating that end-to-end deep learning techniques yield models that perform at state-of-the-art levels, even using a relatively small dataset, (2) assessing the effects of model specification and data augmentation methods on mammography classification performance, and (3) showing that the use of network visualization techniques drawn from computer vision enables the relation of convolutional features to clinically relevant phenomena in a way that can both increase confidence in the outputs of CNN-based CAD outputs while maintaining the potential scientific discovery.

\section{Methods and Data}\label{sec:Methods}

\subsection{Data and Preprocessing}
\begin{figure}[htb]
	\centering
	\includegraphics[width=0.5\textwidth]{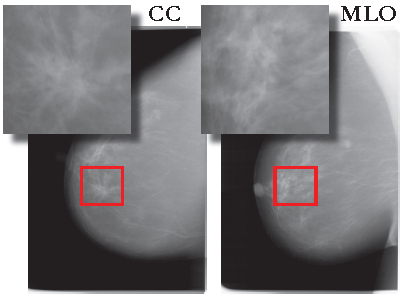}
	\caption{\textbf{Example of DDSM Data.}  \small{Example CC and MLO views of a mammogram.  The red square represents the crop of the image based off of the radiologist-annotated seed pixel.}}
	\label{fig:data}
\end{figure}

The data for this study were a subset of 2085 digital mammography images from a total of 1092 patients from the publicly available Digital Database for Screening Mammography (DDSM)\citep{DDSM}.  53\% of the data are benign cases while the rest are malignant.  Ground truth for this subset of the DDSM dataset was provided by the DDSM resource, typically by biopsy of lesions. These images fall into four classes: normal, cancer, benign, and benign without callback.  Normal cases represent an image for a previous exam for a patient with a normal screening exam four years later.  Benign images represent screenings that caused enough concern to require investigation by another method (ultrasound, pathology, etc.), but ultimately end up being non-cancerous.  Similarly, benign without callback indicates that a physician marked some aspect of the case as worth tracking, but did not require further followup.  Cancer cases were those confirmed to be malignant cases of breast cancer.  Each study contains two separate images, the standard mediolateral oblique (MLO) and craniocaudal (CC) views.  Examples of each can be found in Fig. \ref{fig:data}.  Importantly, MLO and CC views are two different projections that cannot be co-registered, and thus represent different pieces of information about the subject. This dataset is of relatively small size in the context of deep learning and computer vision, but remains widely used in the mammography literature \cite{heath1998current}, thus representing a useful benchmark for comparison to the methods presented here.

\indent The only non-automatic preprocessing step we commit is to crop mammograms to a $750 \times 750$ pixel region around a radiologist-provided seed pixel, demonstrated by the inset figures of Fig. \ref{fig:data}).  We then center and scale the images via mean-variance normalization to achieve image values that fall generally between -3 and 3 in magnitude.  Additionally, given that images are often of slightly different size, each image is resized to $256 \times 256$ and a random $224 \times 224$ crop is obtained to align with the standard input dimensionality requirements of most classic deep learning architectures.  This crop is randomly generated in training, and deterministically generated for testing (at the center of the image).  The images in the training set are also randomly rotated by an integer number of degrees to enable useful dataset augmentation. Training and test sets are created via random splitting, with the test set comprising a random selection of 10\% of the full DDSM dataset and the validation set comprising 10\% of the aforementioned training set.

\subsection{Multi-Modal Convolutional Neural Network}
We investigate two deep CNNs that are based on architectures that have been successful in general computer vision tasks, but that contain several key architectural changes: (1) a multi-modal network with two independent convolutional branches and (2) a network that has both MLO and CC images ``vote" on the final prediction scores, but share all weights in between.

\begin{figure}[htb]
	\centering
	\begin{subfigure}[h]{0.45\textwidth}
		\includegraphics[width=\textwidth]{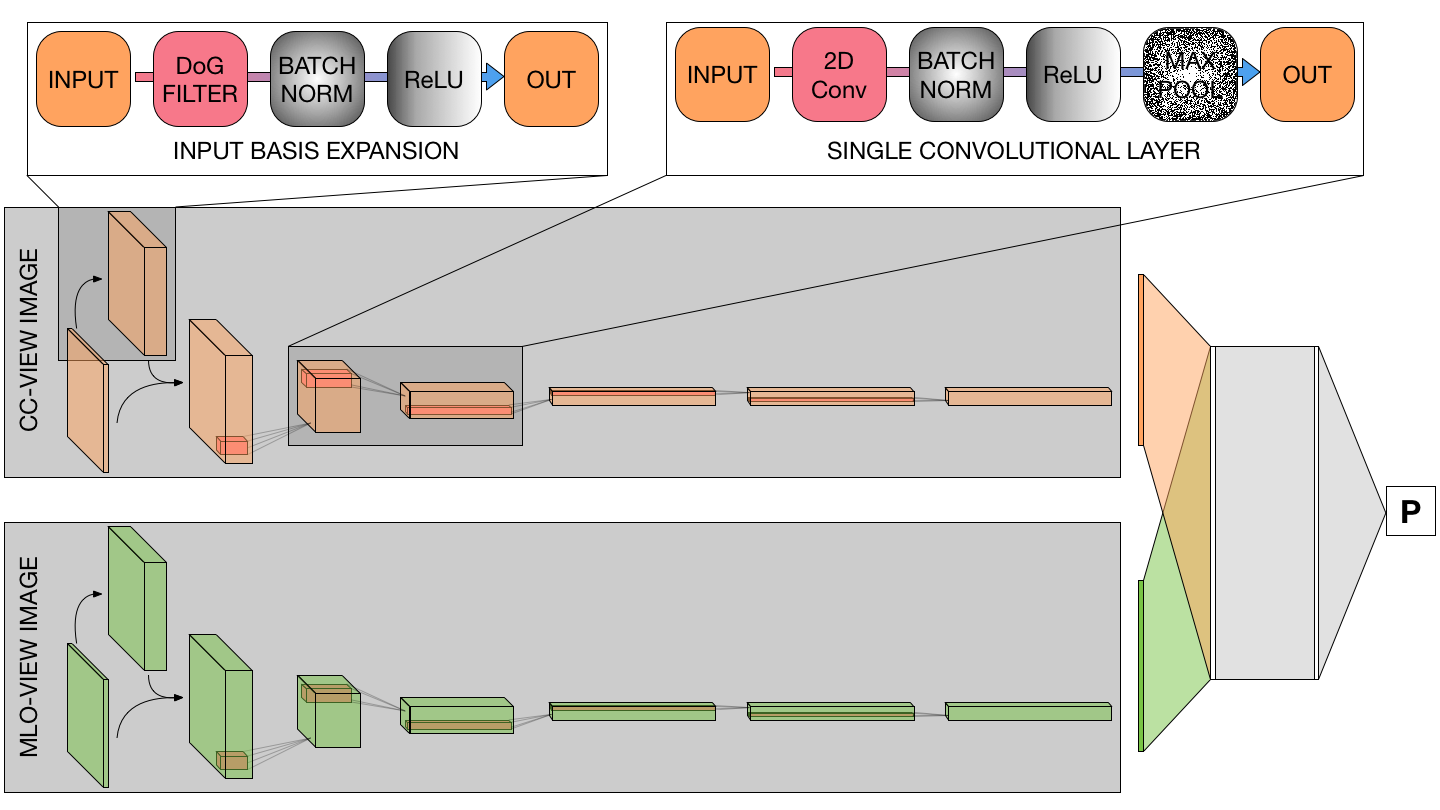}
		\caption{Multi-modal Training}
	\end{subfigure}
	~
	\begin{subfigure}[h]{0.45\textwidth}
		\includegraphics[width=\textwidth]{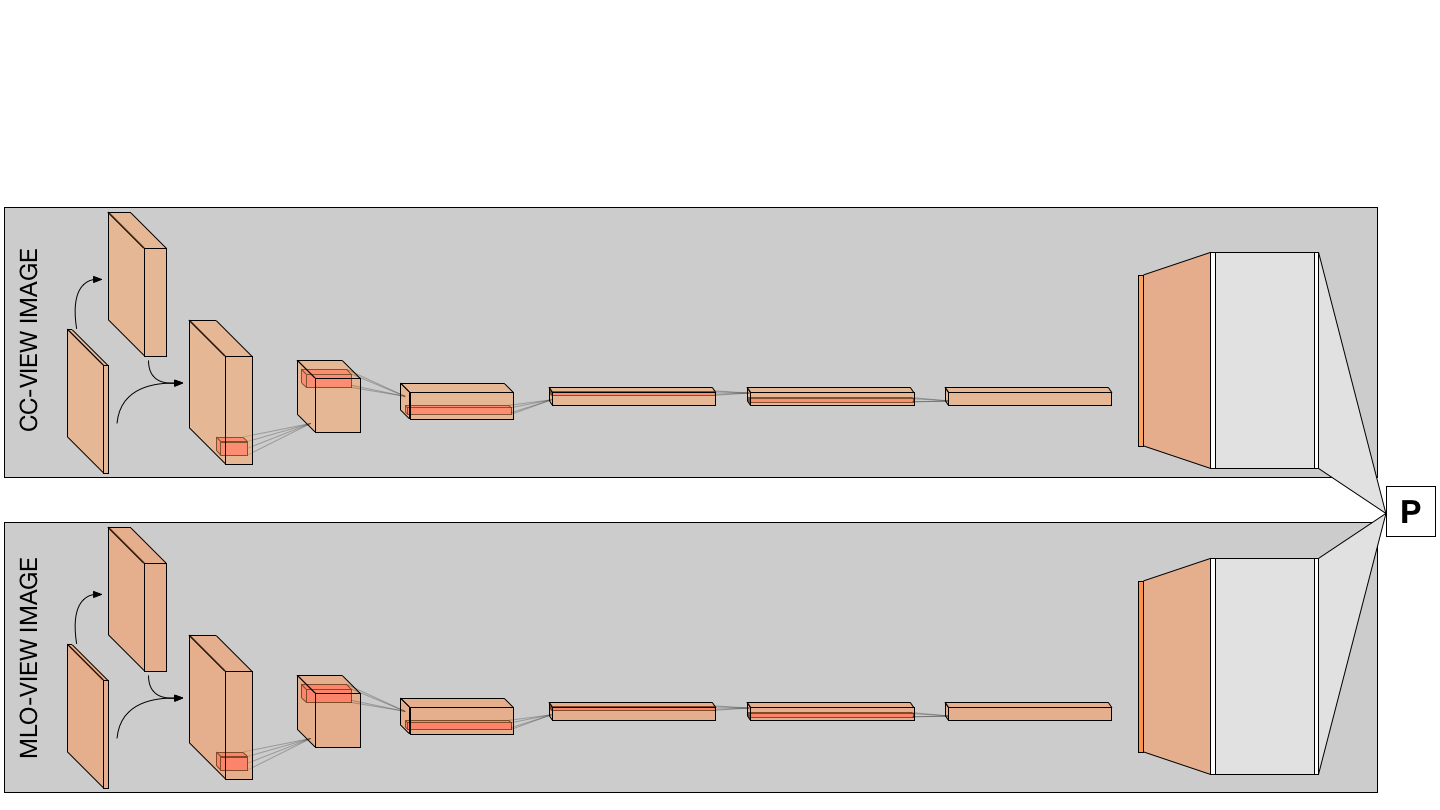}
		\caption{Parallel Training}
	\end{subfigure}
	\caption{\textbf{Algorithm Design.} \small{(a) Two convolutional branches with distinct weights will feed in to the same fully connected network.  (b) Weights for the MLO and CC networks are essentially shared since a single network is tasked with learning features from both views.}}
	\label{fig:algorithm}
\end{figure}

\indent We define the architecture seen in Fig. \ref{fig:algorithm}a as multimodal training, a network that creates two completely independent set of convolutional layers; one will take input from the CC images and the other will take input from the MLO images.  At the end of the convolutional layers, we concatenate the flattened feature vectors and push them through a fully connected network.  An important characteristic of this model, for instance, is the fact that it was designed to handle multiple images describing the same subject, but from different spatial viewpoints.  Given the geometric difference between the MLO and CC views, it is not possible to co-register the images in a useful way.  Further, treating these two views as separate channels in the input layer of a CNN would not necessarily be appropriate, as a fundamental advantage of CNNs is the fact that application of a convolutional filter to a set of pixels or higher-level activations fundamentally incorporates true spatial locality within the field of view of the kernel.  In doing so, we are able to extract key features from each image without cross-contamination during filter optimization while still retaining insight from the combined information in both views via integration at higher levels.  Intuitively, we want to look at the images separately but think about them together.  As seen in Fig. \ref{fig:algorithm}b, we also train a single network to do both tasks of interpreting CC and MLO images.  However, during test time, we average the softmax probability scores produced by the CC branch and the MLO branch to make sure we incorporate information from both the CC and MLO views.  MLO and CC images are trained independently using the same network, but both images vote on the final prediction together.  Though this model may make less sense from a clinical perspective, it has the mathematical advantage of having approximately half the degrees of freedom but effectively twice the amount of data.  Also, should there be any shared image features between CC and MLO images (as there should be), it would be easier for the network to just learn from both of these images in parallel rather than having to learn them independently in both branches of \ref{fig:algorithm}a.

\indent The core algorithm shown in Fig. \ref{fig:algorithm} is the famous AlexNet \cite{krizhevsky2012imagenet}.  We also replace this model with several successful architectures (see Table \ref{table:Comparison}), for comparison.  The only modification we make is a preprocessing layer: 8 scales of Difference of Gaussian filters (Eq. \ref{eq:DoG}) at $\sigma = \{ \sqrt{2}, 2, 2\sqrt{2}, 4, 4\sqrt{2}, 8, 8\sqrt{2}, 16 \}$.  We set these 8 filters to be constant and attach them to the computational graph such that the original image passes through the filter bank of 8.  The result (now dimension $224\times 224 \times 8$) is concatenated with the original grayscale image and fed as an input to our convolutional neural network (which now takes an input of 9 channels).  We define our Difference of Gaussian filters as

\begin{equation}\label{eq:DoG}
\text{DoG} (\sigma) = \frac{1}{2 \pi \sigma^2} e^{-\frac{x^2 + y^2}{2 \sigma^2}} - \frac{1}{\pi \sigma^2} e^{-\frac{x^2 + y^2}{\sigma^2}},
\end{equation}
where DoG is our Difference of Gaussian filter, $\sigma$ is the scale of our filter (in pixels), and $x$ and $y$ are coordinates from the center of our filter (in pixels).

\indent The output layer is a standard softmax classifier with regularization via both direct L2 norm regularization and a dropout layer. Finally, an Adam optimizer is used for training the network via batch stochastic gradient descent.

\subsection{Visualization}

A key question in applying CNNs to medical imagery is whether the features learned via mathematical optimization reflect clinically relevant phenomena.  While in many image classification contexts low-level features can often be interpreted as standard constructs such as edge detection kernels, extraction of higher-level features may be required to demonstrate that features learned by a CNN reflect attributes of known clinical or diagnostic importance.  To investigate this question in the present study, we utilize an adaptation of the DeepDream algorithm \cite{mordvintsev2015inceptionism} to accentuate characteristics assigned by the network to each class in the DDSM dataset (in this case, simply benign or malignant), seen in algorithm 1.

\indent We use a trained CNN in addition to stochastic data modification procedures to modify an original input image to reflect an input image that the CNN would be most likely to classify as either malignant or benign.  Note that the data augmentation procedure involves randomly flipping and rotating the image, in addition to zeroing out boxes of a random size within the image field to ensure that all areas of the image are contributing to the Directed Dream output, and that no single part of the image is having a dominant effect on the Directed Dream process.

\begin{center}
\begin{minipage}[c]{0.9\textwidth}
\noindent\makebox[\linewidth]{\rule{\textwidth}{0.4pt}}
\noindent \textbf{Algorithm 1 Directed Dream.}  We define some objective function $f:x\mapsto y$ where $x$ is the input image and $y$ is a class score based on two logit scores ($S^{(0)}$ and $S^{(1)}$).
\noindent\makebox[\linewidth]{\rule{\textwidth}{0.2pt}}
\begin{enumerate}[noitemsep]
\item Train a valid neural network: $NN$
\item \texttt{x} $\leftarrow$ \{$I_{\text{CC View}}, I_{\text{MLO View}}$\} \hfill 
\item \texttt{for i in 1:MaxIter} \hfill
\item \hspace*{1em} Perform forward stochastic alteration of \texttt{x} 
\item \hspace*{1em} Feed \texttt{x} forward through $NN$
\item \hspace*{1em} Compute gradient $\frac{\partial y}{\partial x}$ as $\left[\frac{\partial y}{\partial S^{(0)}}, \frac{\partial y}{\partial S^{(1)}}\right] = [\pm 1, \mp 1] $
\item \hspace*{1em} Z-score normalize $\frac{\partial y}{\partial x}$, clipping values at $[-0.05, 0.05]$
\item \hspace*{1em} \texttt{x := x +} $\frac{\partial y}{\partial x}$ * \texttt{learning\_rate} 
\item \hspace*{1em} Clip \texttt{x} at $[-3,3]$
\item \hspace*{1em} Perform reverse stochastic alteration of \texttt{x}
\end{enumerate}
\noindent\makebox[\linewidth]{\rule{\textwidth}{0.2pt}}
\end{minipage}
\end{center}

\subsection{Non-Deep Learning Comparisons}

Finally, we recreate methods from two non-deep learning approaches that focused on using texture features, for added comparison.  Sahiner's method involves performing a rubber-band straightening transform (RBST) on the image, converting the margin of the image into a straight line.  After this, texture features, morphological features, and run-length statistics are used to extract a feature vector.  These feature vectors were used in a linear discriminant analysis model with feature selection for classification.  \cite{sahiner1998computerized}  Huo's method extracts a feature vector using five main features: spiculation measure, sharpness, average gray level, contrast, and texture.  These feature vectors were used in a hybrid rule-based neural network classifier.\cite{huo1998automated}  These methods were recreated for means of comparison, and all reported results are on the same subset of DDSM data used for the neural network based methods.

\subsection{Evaluation}

For all results presented, training and testing was done on the subset of DDSM cases described above.  For the neural network based models seen in table \ref{table:Comparison}, all results are tested on a held-out test set comprising 100 patients ($\sim$10\% of the total data).  To help select our model and choose hyperparameters, we used a validation set of 99 patients and a training set of 893 patients.  These sets were held constant for all the neural networks tested against the ground truth provided by the DDSM resource.  The two non-deep learning methods were tested on a held-out test set of 206 patients.

\section{Results and Discussion}\label{sec:Results}

\begin{figure}[htb]
	\centering
	\begin{subfigure}[h]{0.45\textwidth}
		\includegraphics[width=\textwidth]{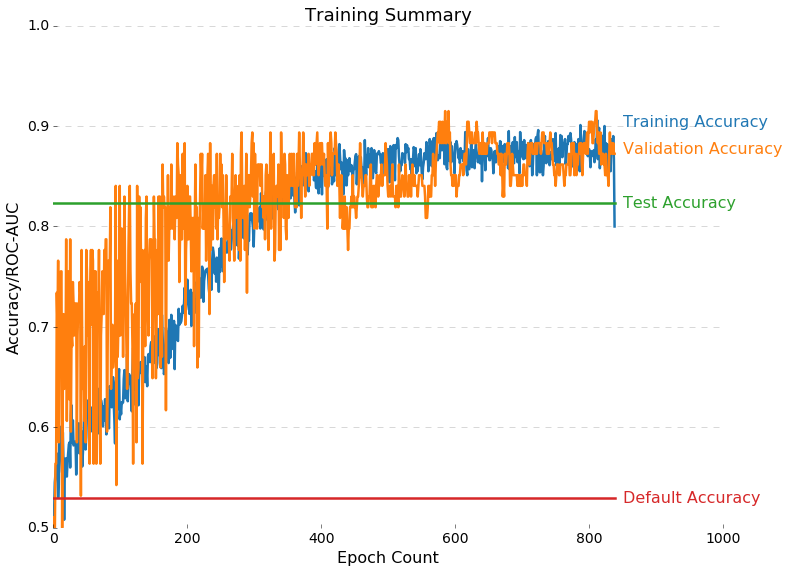}
		\caption{Training/Validation Curve}
	\end{subfigure}
	~
	\begin{subfigure}[h]{0.45\textwidth}
		\includegraphics[width=\textwidth]{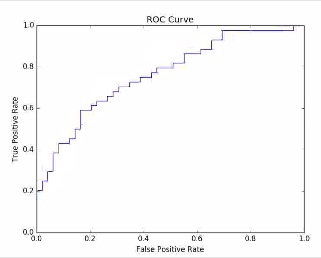}
		\caption{ROC Curve}
	\end{subfigure}
	\caption{\textbf{Training Results.} \small{(a) A curve showing the training, validation, and test-time results for a selected GoogLe Net model (within the 100 ensemble).  This one achieves a held-out test set accuracy of 0.84.  (b) A ROC curve for our model on the held-out test set.  We achieve an ROC AUC of 0.85.}}
	\label{fig:results}
\end{figure}

We use the process described above to perform the classification task for the DDSM mammography data.  Key hyperparameters for the final model are a dropout probability of 0.1, a learning rate of 0.001 with decay rate of 0.99 per epoch, and a regularization coefficient of 0.000001.  Model training was performed on four Titan X (Pascal) GPU's in parallel, and took 4 hours for training over 800 epochs with a batch size of 120.  \textbf{The best model reported here was an ensembled GoogLeNet-based architecture (in parallel training) that was able to achieve a held-out test accuracy of 85\%,} far above the majority classifier accuracy of 53\%.  As shown in Fig. \ref{fig:results}, we observe a well-conditioned training process that does not show signs of overfitting.  The right panel of Fig. \ref{fig:results} illustrates the ROC curve for the model at one particular point during the training process. 

\begin{table}[h]
 \caption{\textbf{Performance Comparison.} \small{We compare different performances of different algorithms on the task of benign/malignant classification on this subset of DDSM.}}\label{table:Comparison}
\begin{center}
\begin{tabular}{L{2.4cm}|C{0.8cm}C{0.8cm}C{0.8cm}|C{0.8cm}C{0.8cm}C{0.8cm}|C{0.8cm}C{0.8cm}C{0.8cm}}
\toprule[1.5pt]
Method&\multicolumn{3}{l|}{Multi-modal w/ Aug}&\multicolumn{3}{l|}{Parallel w/o Aug}&\multicolumn{3}{l}{Parallel w/ Aug}\\
&Acc.&AUC&Loss&Acc.&AUC&Loss&Acc.&AUC&Loss\\
\midrule
\midrule
AlexNet \cite{krizhevsky2012imagenet} & 0.58 & 0.64 & 1.18 & 0.70 & 0.82 & 1.24 & 0.70 & 0.78 & 0.65  \\
VGG16 \cite{simonyan2014very} & - & - & -& 0.70 & 0.79 & 1.20 & 0.72 & 0.88 & 0.58 \\
GoogLe Net \cite{szegedy2015going} & 0.77 & 0.78 & 1.94& 0.78 & 0.87 & 1.51 & 0.78 & 0.87 & 0.53  \\
InceptionV3 \cite{szegedy2016rethinking}& 0.73 & 0.75 & 2.10& 0.74 & 0.84 & 1.91 & 0.84 & 0.87 & 1.49 \\
Residual Net \cite{he2016deep} & 0.71 & 0.74 & 2.93& 0.72 & 0.80 & 0.78 & 0.74 & 0.81 & 0.64  \\
\midrule
\textbf{GoogLe Net} & \multicolumn{6}{l|}{\textbf{Ensemble of 100 Parallel Networks}} & \textbf{0.85} & \textbf{0.91} & \textbf{0.45} \\
\midrule
Sahiner et. al.\cite{sahiner1998computerized} & \multicolumn{9}{l}{Texture Feature Analysis with Accuracy of 0.71}\\
Huo et. al.\cite{huo1998automated} & \multicolumn{9}{l}{Texture Feature Analysis with Accuracy of 0.60}\\
\midrule
\midrule
\end{tabular}
\end{center}
\end{table}

\indent We can see from our results in Table \ref{table:Comparison} that augmentation generally improves performance.  This is not too much of a surprise.  However, we find that a parallel implementation works better than a multi-modal one.  This is most likely due to the fact that our networks have a high level of variance error while already being quite unbiased.  Thus, increasing the parameter size and pushing the CC and MLO views through different convolutional branches does not confer significant added benefit.  Instead, it is more helpful to reinforce the shared features by essentially doubling our data with the parallel implementation as the network probably has enough degrees of freedom to learn distinct features for both views.  Finally, we took our best performing network (as defined by test time loss) and trained 100 networks to create a saturated ensemble, which gave us our best performance.  This not only does the best in terms of accuracy, ROC AUC, and loss, but is at least in the same tier as other state of the art non-deep learning algorithms.

\begin{figure}[htb]
	\centering
	\begin{subfigure}[h]{0.48\textwidth}
        \includemovie[text={\includegraphics[width=\textwidth]{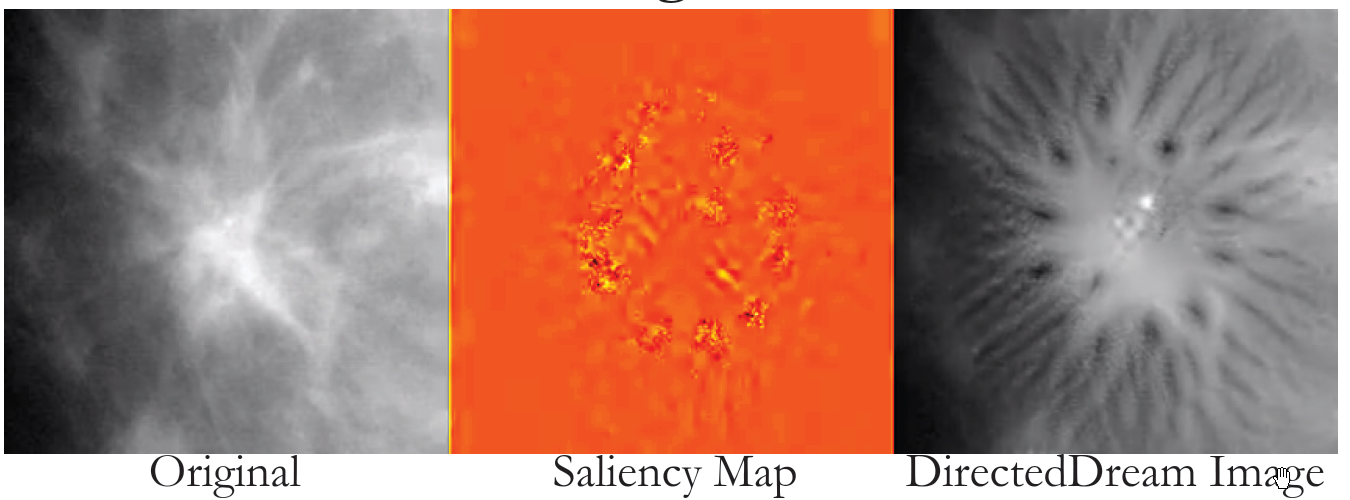}}]{\textwidth}{}{malignant_vid.mp4}
		\caption{Malignant Visualization}
	\end{subfigure}
	~
	\begin{subfigure}[h]{0.48\textwidth}
		\includemovie[text={\includegraphics[width=\textwidth]{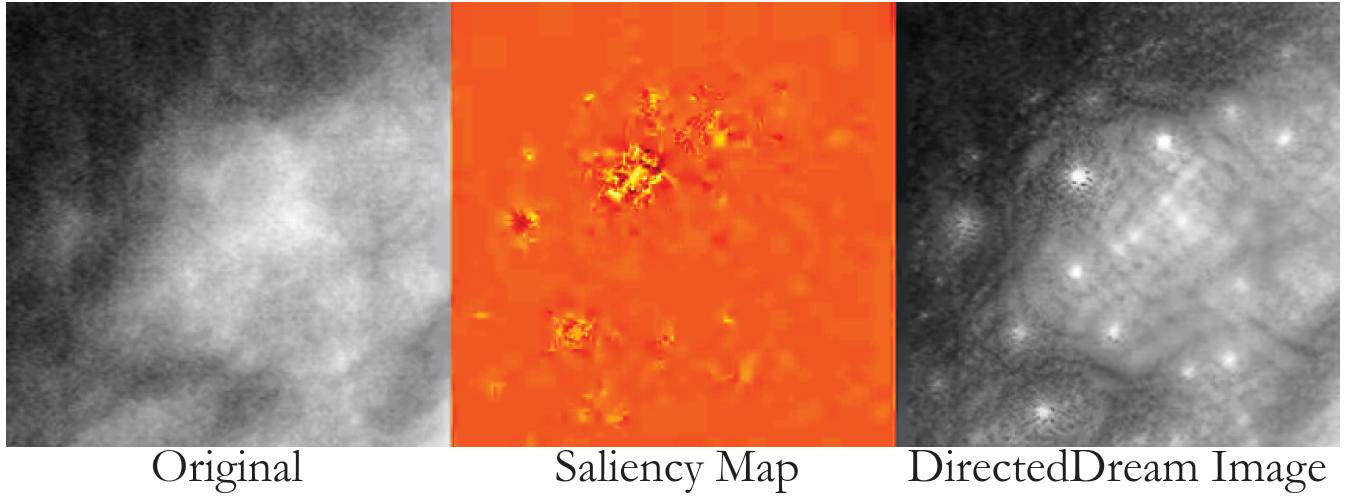}}]{\textwidth}{}{benign_vid.mp4}
		\caption{Benign Visualization}
	\end{subfigure}
	\caption{\textbf{Visualization.} \small{Each image is split into three sections: the original input image, the most recent saliency map added in the stochastic gradient ascent cycle, and the directed dream visualization image.  Note that this figure is a video with each frame a step in the iterative Directed Dream algorithm.}}
	\label{fig:visualization}
\end{figure}

\indent Once this network has been trained, we create our Directed Dream images that maximize both the "malignant" class score and the "benign" class score.  As shown in Fig. \ref{fig:visualization}, the Directed Dream algorithm takes the original image and repeatedly applies stochastic gradient ascent updates commensurate with saliency maps such as the ones displayed.  While a highly-connected starburst-like pattern is formed in the Directed Dream image for the malignant case, we see a ``starry night" effect in the benign case, wherein areas of relative brightness in the original image are consolidated.  These visualizations are CNN-based hallucinations that create features that would maximize a certain class's score.  Thus, certain aspects of the image will be heightened and exaggerated, but other features may be completely fabricated.  However, by analyzing these ``hallucinations,'' we can strive to understand the type of features that our network extracts.  The DirectedDream algorithm will create patterns that correspond to high malignant/benign score based on images seen in the training data.

\indent The malignant visualization (Fig. \ref{fig:visualization}a), presents an interesting radiographic pattern. The mass is bordered with elongated spikes emanating from the tumor edges in a starburst pattern, known in radiology as spiculation.  Tumor infiltration and desmoplastic reaction cause a spiculated pattern in histology and mammography \cite{franquet1993spiculated}. The spiculated soft tissue mass is the most specific mammographic feature of breast cancer; nearly 90 \% of these lesions represent invasive breast cancer \cite{esserman2012diagnostic}.  In contrast, the benign image shows almost round legion like features, minimizing the starburst like connection between any two cluster of bright pixels.  We can see clearly that the two transformations for benign and malignant lesions are opposite sides of the same coin, where one transformation maximizes the gradient features that the other minimizes.

\indent Furthermore, from the central block in Fig. \ref{fig:visualization} that shows the saliency map, we can see that a single saliency map alone is not enough to fully demonstrate the higher level abstractions that the network detects.  Playing the video in Fig. \ref{fig:visualization}, we can see that at each iteration of the Directed Dream algorithm, the saliency map seems noisy and uninterpretable.  However, when the saliency maps are collapsed onto each other on top of the original image, they form a neater, more-interpretable image.

\indent The close topological agreement between radiological spiculation and the malignant Direct Dream output demonstrates that this network has learned a clinically relevant representation of malignancy.  Such a demonstration of clinically relevant feature uptake by a deep CNN has not commonly been shown, and the implication that the CNN is not only performing well on a classification task, but is also making judgments based on features similar to those used by experts, represents a significant piece of evidence validating the use of these architectures in automated radiological applications.

\section{Conclusions}

We have shown a comprehensive study of neural network designs for the task of malignant/benign classification for breast mammography.  With a dataset of $\sim$1000 patients, it seems best to minimize the variance error created by the network by using a compact network (like GoogLe Net).  Also, we can see that a single network tasked with understanding both CC and MLO views can outperform a network that has an architecture built to process both independently.  This probably stems from the fact that there is enough degrees of freedom within any of these neural networks to easily learn these two tasks while giving a single network multiple tasks can possibly help it share relevant features between CC and MLO processing.

\indent Furthermore, through our novel visualization method, we can start to unpack the black box of our neural network.  We see that the CNN learns clinically relevant features, giving credibility to our results.  By framing our visualization method in an iterative fashion that evolves the image over time, we are able to see higher levels of abstraction in the features our network has learned in comparison to one-time gradient methods such as saliency maps.

\subsubsection*{Acknowledgments}

This work was supported in part by grants from the National Cancer Institute, National Institutes of Health, U01CA142555, 1U01CA190214, and 1U01CA187947.

\bibliographystyle{plain}
\bibliography{DDSM_bib}

\begin{thebibliography}{10}

\bibitem{bekker2016}
Alan Bekker and et.al.
\newblock Multi-view probabilistic classification of breast
  microcalcifications.
\newblock {\em IEEE Transactions On Medical Imaging}, 35(2):645--653, 2016.

\bibitem{chu2015}
Jinghui Chu, Hang Min, Li~Liu, and Wei Lu.
\newblock A novel computer aided breast mass detection scheme based on
  morphological enhancement and slic superpixel segmentation.
\newblock {\em Medical Physics}, 42(7):3859--3869, 2015.

\bibitem{DDSM}
DDSM.
\newblock Usf digital mammography home page.
\newblock November 2016.

\bibitem{de2016machine}
Marleen de~Bruijne.
\newblock Machine learning approaches in medical image analysis: From detection
  to diagnosis.
\newblock {\em Medical Image Analysis}, 33:94--97, 2016.

\bibitem{esserman2012diagnostic}
Laura~J Esserman and Bonnie~N Joe.
\newblock Diagnostic evaluation of women with suspected breast cancer.
\newblock {\em Up To Date.}, 2012.

\bibitem{franquet1993spiculated}
Tomas Franquet and et.al.
\newblock Spiculated lesions of the breast: mammographic-pathologic
  correlation.
\newblock {\em Radiographics}, 13(4):841--852, 1993.

\bibitem{giger2008anniversary}
Maryellen~L Giger, Heang-Ping Chan, and John Boone.
\newblock Anniversary paper: history and status of cad and quantitative image
  analysis: the role of medical physics and aapm.
\newblock {\em Medical physics}, 35(12):5799--5820, 2008.

\bibitem{gutman2013mr}
David~A Gutman and et.al.
\newblock Mr imaging predictors of molecular profile and survival:
  multi-institutional study of the tcga glioblastoma data set.
\newblock {\em Radiology}, 267(2):560--569, 2013.

\bibitem{he2016deep}
Kaiming He, Xiangyu Zhang, Shaoqing Ren, and Jian Sun.
\newblock Deep residual learning for image recognition.
\newblock pages 770--778, 2016.

\bibitem{heath1998current}
Michael Heath, Kevin Bowyer, Daniel Kopans, P~Kegelmeyer~Jr, Richard Moore,
  Kyong Chang, and S~Munishkumaran.
\newblock Current status of the digital database for screening mammography.
\newblock In {\em Digital mammography}, pages 457--460. Springer, 1998.

\bibitem{huo1998automated}
Zhimin Huo and et.al.
\newblock Automated computerized classification of malignant and benign masses
  on digitized mammograms.
\newblock {\em Academic Radiology}, 5(3):155--168, 1998.

\bibitem{kooi2017large}
Thijs Kooi and et.al.
\newblock Large scale deep learning for computer aided detection of
  mammographic lesions.
\newblock {\em Medical image analysis}, 35:303--312, 2017.

\bibitem{krizhevsky2012imagenet}
Alex Krizhevsky and et.al.
\newblock Imagenet classification with deep convolutional neural networks.
\newblock 2012.

\bibitem{mordvintsev2015inceptionism}
Alexander Mordvintsev, Christopher Olah, and Mike Tyka.
\newblock Inceptionism: Going deeper into neural networks.
\newblock {\em Google Research Blog. Retrieved June}, 20, 2015.

\bibitem{rao2010widely}
Vijay~M Rao, David~C Levin, Laurence Parker, Barbara Cavanaugh, Andrea~J
  Frangos, and Jonathan~H Sunshine.
\newblock How widely is computer-aided detection used in screening and
  diagnostic mammography?
\newblock {\em Journal of the American College of Radiology}, 7(10):802--805,
  2010.

\bibitem{sahiner1998computerized}
Berkman Sahiner, Heang-Ping Chan, Nicholas Petrick, Mark~A Helvie, and
  Mitchell~M Goodsitt.
\newblock Computerized characterization of masses on mammograms: The rubber
  band straightening transform and texture analysis.
\newblock {\em Medical Physics}, 25(4):516--526, 1998.

\bibitem{simonyan2013deep}
Karen Simonyan, Andrea Vedaldi, and Andrew Zisserman.
\newblock Deep inside convolutional networks: Visualising image classification
  models and saliency maps.
\newblock {\em arXiv preprint arXiv:1312.6034}, 2013.

\bibitem{simonyan2014very}
Karen Simonyan and Andrew Zisserman.
\newblock Very deep convolutional networks for large-scale image recognition.
\newblock {\em arXiv preprint arXiv:1409.1556}, 2014.

\bibitem{szegedy2016rethinking}
Christian Szegedy and et.al.
\newblock Rethinking the inception architecture for computer vision.
\newblock pages 2818--2826, 2016.

\bibitem{szegedy2015going}
Christian Szegedy, Wei Liu, Yangqing Jia, Pierre Sermanet, Scott Reed, Dragomir
  Anguelov, Dumitru Erhan, Vincent Vanhoucke, and Andrew Rabinovich.
\newblock Going deeper with convolutions.
\newblock pages 1--9, 2015.

\bibitem{wang2016deep}
Dayong Wang, Aditya Khosla, Rishab Gargeya, Humayun Irshad, and Andrew~H Beck.
\newblock Deep learning for identifying metastatic breast cancer.
\newblock {\em arXiv preprint arXiv:1606.05718}, 2016.

\bibitem{wang2012machine}
Shijun Wang and Ronald~M Summers.
\newblock Machine learning and radiology.
\newblock {\em Medical image analysis}, 16(5):933--951, 2012.

\end{thebibliography}

\end{document}